\documentclass[runningheads]{llncs}
\usepackage{graphicx}
\usepackage{bm}
\usepackage{mathtools}
\usepackage{amsfonts}
\usepackage{caption}
\usepackage{subcaption}
\usepackage{soul}

\usepackage[colorinlistoftodos]{todonotes}

\usepackage[colorlinks=true]{hyperref}

\begin{document}
\title{InfoMask: Masked Variational Latent Representation to Localize Chest Disease}
\titlerunning{InfoMask: Disease Localization from Chest X-rays}
%
\author{Saeid Asgari Taghanaki\inst{1,2,3} \and
Mohammad Havaei\inst{2} \and
Tess Berthier\inst{2} \and
Francis Dutil\inst{2} \and
Lisa Di Jorio\inst{2} \and
Ghassan Hamarneh\inst{3} \and
Yoshua Bengio\inst{1}}

\authorrunning{Asagri Taghanaki et al.}
%
\institute{MILA, Université de Montréal, Canada \and
Imagia Inc., Montreal, Canada \\
\and
School of Computing Science, Simon Fraser University, Canada}
%
\maketitle              
\begin{abstract}
The scarcity of richly annotated medical images is limiting supervised deep learning based solutions to medical image analysis tasks, such as localizing discriminatory radiomic disease signatures. Therefore, it is desirable to leverage unsupervised and weakly supervised models. Most recent weakly supervised localization methods apply attention maps or region proposals in a multiple instance learning formulation. While attention maps can be noisy, leading to erroneously highlighted regions, it is not simple to decide on an optimal window/bag size for multiple instance learning approaches. In this paper, we propose a learned spatial masking mechanism to filter out irrelevant background signals from attention maps. The proposed method minimizes mutual information between a masked variational representation and the input while maximizing the information between the masked representation and class labels. This results in more accurate localization of discriminatory regions. We tested the proposed model on the ChestX-ray8 dataset to localize pneumonia from chest X-ray images without using any pixel-level or bounding-box annotations.

\end{abstract}
\section{Introduction}
Around 50,000 deaths attributed to pneumonia are reported every year in the US alone~\cite{CDC}. Chest X-rays are currently the most adopted imaging modality for detecting pneumonia~\cite{who1}. The large increase in imaging studies, scarcity of radiologists and associated expense and intra-/inter-rater variability, has resulted in an acceleration in the development and adoption of automated image-based disease classification methods. In the last decade, deep learning methods have resulted in great successes in classification of natural and medical images. Locating discriminatory regions in images, along with the predicted class, renders deep models' decisions more interpretable and trustworthy. Localization of disease-indicator regions (i.e., radiomic biosignatures) is particularly important for medical applications since it reveals whether the machine diagnosis was based on the presence/absence of disease and not biased towards some unique yet unintuitive and unrelated pattern that happens to be exhibited among the training examples.

\subsection{Related Work}

The past few years have witnessed numerous advances in deep learning methods for localizing objects and detecting discriminatory regions in images. 

\noindent\textbf{Multiple instance learning and region-based methods.} Bency et al. \cite{bency2016weakly} and Teh et al. \cite{teh2016attention} applied region-proposal and beam search based methods to localize objects from natural (i.e., non-medical) images. Training such hybrid localization-classification models requires large amounts of bounding-box level image annotations, which can suffer from rater-variability and can be prohibitively expensive or time consuming. Several existing methods~\cite{bilen2016weakly,kumar2010self,song2014learning,song2014weakly,wang2014weakly} formulate the weakly-supervised localization as a multiple instance learning (MIL) problem. However, like for region-proposal based methods, it is difficult to find an optimal window size. 

\noindent \textbf{Attention/activation based methods.} Similarly to previous works~\cite{yan2018weakly,rajpurkar2017chexnet,wang2018weakly}, Wei et al.~\cite{wei2018ts2c} proposed an activation map based framework to produce tight bounding boxes around objects. However, in the context of object localization, there might be erroneously detected regions (false positives) or regions/activations which spread over unrealistically wide ranges. This is because saliency maps~\cite{simonyan2013deep} are usually noisy.

Several works have attempted to smooth or regularize the saliency or attention maps~\cite{smilkov2017smoothgrad} and prevent important features to be neglected due to saturation~\cite{ancona2018towards,shrikumar2017learning}. To produce class-discriminative `explanation maps', the gradient-weighted class activation mapping method GradCAM \cite{SelvarajuDVCPB16} was used to capture the importance of a particular features of a target class. However, GradCAM and similar approaches are applied \textit{after} a model is trained, i.e., there is no \textit{explicit} spatial enforcement during training and GradCAM requires class labels during inference. To remove dependency on class labels during test, Fan et al.~\cite{Fan2017AdversarialLN}, trained a masking mechanism simultaneously with a classification network to localize objects. However, their masking mechanism is based on super-pixels which might miss fine details. Zolna et al.~\cite{zolna2018classifier} reformulated the same problem as a min-max game. It is not clear to us how to weigh the regularization term in their proposed loss function and we have concerns about scalability of the method, as it needs to preserve many copies of the model with different parameters to produce different masks using each set of parameters. In this paper, instead of keeping different parameters of a model, we propose to perform variational online mask sampling from a normal distribution using a single model. 

\subsection{Contribution}

The focus of the current study is to develop a method to localize radiologic presentations of pneumonia in novel chest X-ray images while training on data with only image-level labels. The key idea of the proposed approach is to learn a low-dimensional latent parametric probability distribution (regularized by Kullback-Leibler divergence from a standard normal distribution) that encodes the input data and be not only discriminative but also spatially-selective to disregard irrelevant background from input images. To this end, we propose InfoMask, a variational model with a learnt attention mechanism and a sparsity-promoting masking operation.

In this paper we make the following contributions: a) We propose to produce online variational masks during training without the need for class labels during inference; b) we propose a weakly supervised localization method without requiring any choice of window/bag size (which is necessary in competing multiple instance learning formulations); c) we introduce a masking mechanism applied to the latent variational representation to filter non-discriminatory information; and d) we propose minimizing mutual information between the input and latent variational attention maps and increasing the mutual information between the masked latent representation and class labels.

\section{Method}
Given a training set of input images $\mathbf{x}$ and corresponding image-level labels $y$, our goal is to learn the parameters $\bm { \theta }$ of a class-predictive model $\hat{y}=g(\mathbf{x};\bm { \theta })$ that not only has high classification accuracy but also localizes the discriminative regions with minimal inclusion of irrelevant pixels.  The localization is represented via a binary mask $\bm{M}$ and $\bm { \theta }$ is learnt by maximizing $p(y|\bm{x},\bm{M};\bm { \theta })$.


To this end, InfoMask learns to encode a bottleneck random variable $\bm{Z}$ that (i) captures minimal information about the input random variable $\bm{X}$, hence minimizes the encoding of irrelevant information in the input, and (ii) holds maximal information about the distribution of the target label variable $Y$.
Consequently, inspired by Alemi et al.~\cite{alemi2016deep} and the information bottleneck~\cite{Tishby-1999}, we aim at maximizing 
\begin{equation}
   L( \bm { \theta } ) = I ( \bm{Z} , Y ; \bm { \theta } ) - \alpha I ( \bm{Z},\mathbf{X} ; \bm { \theta } ) 
\end{equation}
\noindent where $I(A,B)$ is the mutual information between random variables A and B, and $\alpha$ is a scalar weight.




We model $\bm{z}\sim\mathcal{N}(\bm{\mu_z},\bm{\sigma_z})$ and learn to generate its mean and  variance using  convolutional layers, i.e., 
$\mu_z=f_{e}^{\bm{\mu}}(\bm{x})$ and $\sigma_z=f_{e}^{\bm{\sigma}}(\bm{x})$, and  rewrite $I(\bm{Z},Y;\bm{\theta})$ (and similarly $I(\bm{Z},\bm{X};\bm{\theta})$), for each element of $\bm{Z}$, as: 
\begin{equation}
I(\bm{Z},Y;\bm{\theta},\bm{\mu_z},\bm{\sigma_z})
= \int p(\bm{z},y;\bm{\theta},\bm{\mu_z},\bm{\sigma_z}) 
  \log \frac{p(\bm{z},y;\bm{\theta},\bm{\mu_z},\bm{\sigma_z})} {p(\bm{z};\bm{\theta},\bm{\mu_z},\bm{\sigma_z}) p(y;\bm{\theta })} dx dy
\end{equation}



\begin{figure}[]
\centering
  \includegraphics[width=.8\textwidth]{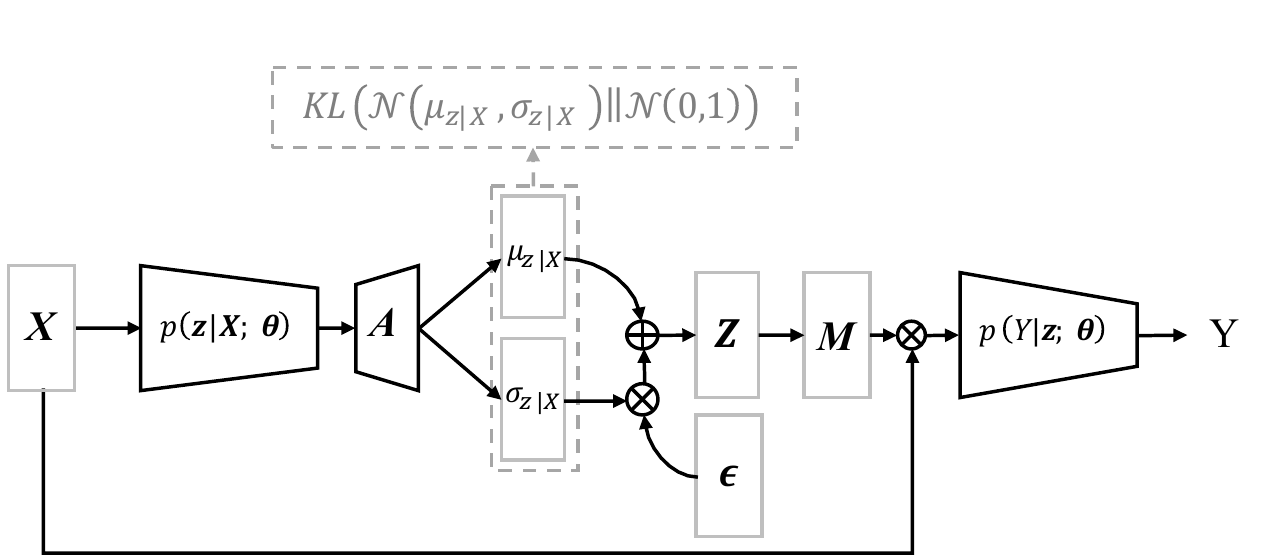}
  \caption{Architecture and components of the proposed model. The input $\bm{X}$ is encoded via $p(\bm{z}|\bm{X};\bm{\theta})$. $\bm{A}$ refers to an attention map computed from the last layer of the encoder using $1\times1$ convolution and ReLU. Note that $\bm{A}$ is upsampled to the size of $\bm{X}$. $\bm{M}$ is the masked latent matrix. $\bm{X}$, $\bm{A}$,  $\bm{\mu}$, $\bm{\sigma}$, $\bm{Z}$, and $\bm{\epsilon}$ are of size $W \times H$.}
  \label{diag1}
\end{figure}

\noindent To sample $\bm{Z}$, we apply the reparameterization trick and write $\bm{z} = f(\bm{x},\epsilon)=\bm{\mu_z} + \bm{\sigma_z} \bm{\epsilon} = \bm{a}_{e}^{\bm{\mu}}(\bm{x}) + \bm{a}_{e}^{\bm{\sigma}}(\bm{x}) \epsilon$, where $\bm{a}_{e}$ is a deterministic function which outputs both $\bm{\mu}$ and $\bm{\sigma}$ and $\bm{\epsilon} \sim \mathcal{N}(0,1)$. We regularize the distribution by penalizing the Kullback-Leibler (KL) divergence from a standard normal distribution.
The final loss function which we aim to minimize is given by
\begin{equation}
    L = \frac { 1 } { N } \sum _ { n = 1 } ^ { N } \mathbb {E } _ { \bm{\epsilon} \sim p ( \epsilon ) } \left[ - \log q \left( y _ { n } | \bm{a} \left( \bm{x} _ { n } , \epsilon \right) \right) \right] + \alpha \mathrm { KL } [ p ( \bm{Z} | \bm{x} _ { n } ) , r ( \bm{Z} ) ]
\end{equation}

\noindent where ${N}$ is the number of training examples, $q(.)$ is the variational approximation function, and $r(\bm{Z})$ is variational approximation. In our case, $\bm{Z}$ is not computed directly from the input but rather by sampling an attention map $\bm{A}$ from which $\mu_z$ and $\sigma_z$ are derived. To \textit{explicitly} enforce the model to generate more focused attention maps, we apply the following masking function $\bm{M}$ with threshold $\tau$ that localizes the discriminative areas of $\bm{Z}$.
\begin{equation}
    \bm{M} =  R \left( \Tilde{\bm{z}} - \tau \right)~~~\text{where}~~~ \ \ \  \Tilde{\bm{z}} =   { (1 + \exp \left( - \bm{z} \right))^{-1} }
\end{equation}
where $R$ is a ReLU function with upper bound of 1, i.e., $R(v)=max(0,min(v,1))$. The block diagram of the proposed method is shown in Fig.~\ref{diag1}.
\vspace*{-5mm}
\vspace*{-10mm}
\section{Data}
For evaluation, we used the NIH ChestX-ray8 Dataset~\cite{wang2017chestx}, which comprises 112,120 X-ray images from 30,805 unique patients with corresponding disease labels. We used 20547, 2568, and 2569 with pneumonia images as train, validation, and test sets, respectively. For training and to evaluate the test classification accuracy, we only used image-level labels. To evaluate the localization performance on test images, we used ground truth bounding boxes manually placed around the diseased areas.

\section{Experiments and Results}
We adopt a simple architecture as a baseline, i.e., an encoder ($ p ( \bm{z} | \bm{X} ; \bm{\theta} )$) of the form [conv(64, 3x3, relu), conv(64, 3x3, relu), maxpooling(2x2), conv(128, 3x3, relu), conv(128, 3x3, relu), maxpooling(2x2), conv(256, 3x3, relu), conv(16, 3x3, relu)] and a classification block ($p ( Y | \bm{z} ; \bm{\theta} )$) of [conv(128, 3x3, relu), maxpooling(2x2), conv(64, 3x3, relu), conv(64, 3x3, relu) maxpooling(2x2), global average pooling, softmax]. 

\begin{figure}
\centering
\begin{subfigure}[b]{.34\linewidth}
\includegraphics[width=\linewidth]{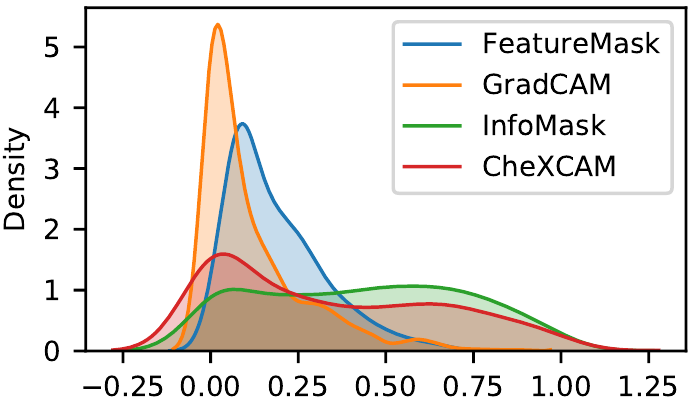}
\caption{IoP}\label{fig:iop}
\end{subfigure}
\begin{subfigure}[b]{.32\linewidth}
\includegraphics[width=\linewidth]{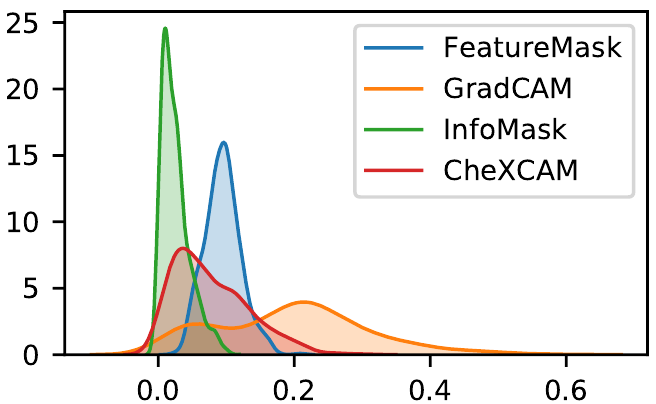}
\caption{FPR}\label{fig:fpr}
\end{subfigure}
\begin{subfigure}[b]{.32\linewidth}
\includegraphics[width=\linewidth]{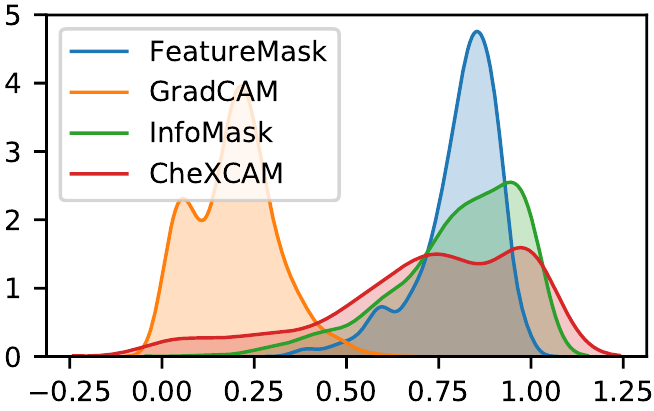}
\caption{FNR}\label{fig:fnr}
\end{subfigure}
\caption{Kernel density estimation of different measures for disease localization}
\label{fig:kde}
\end{figure}

\begin{figure}[]
\centering
  \includegraphics[width=\textwidth]{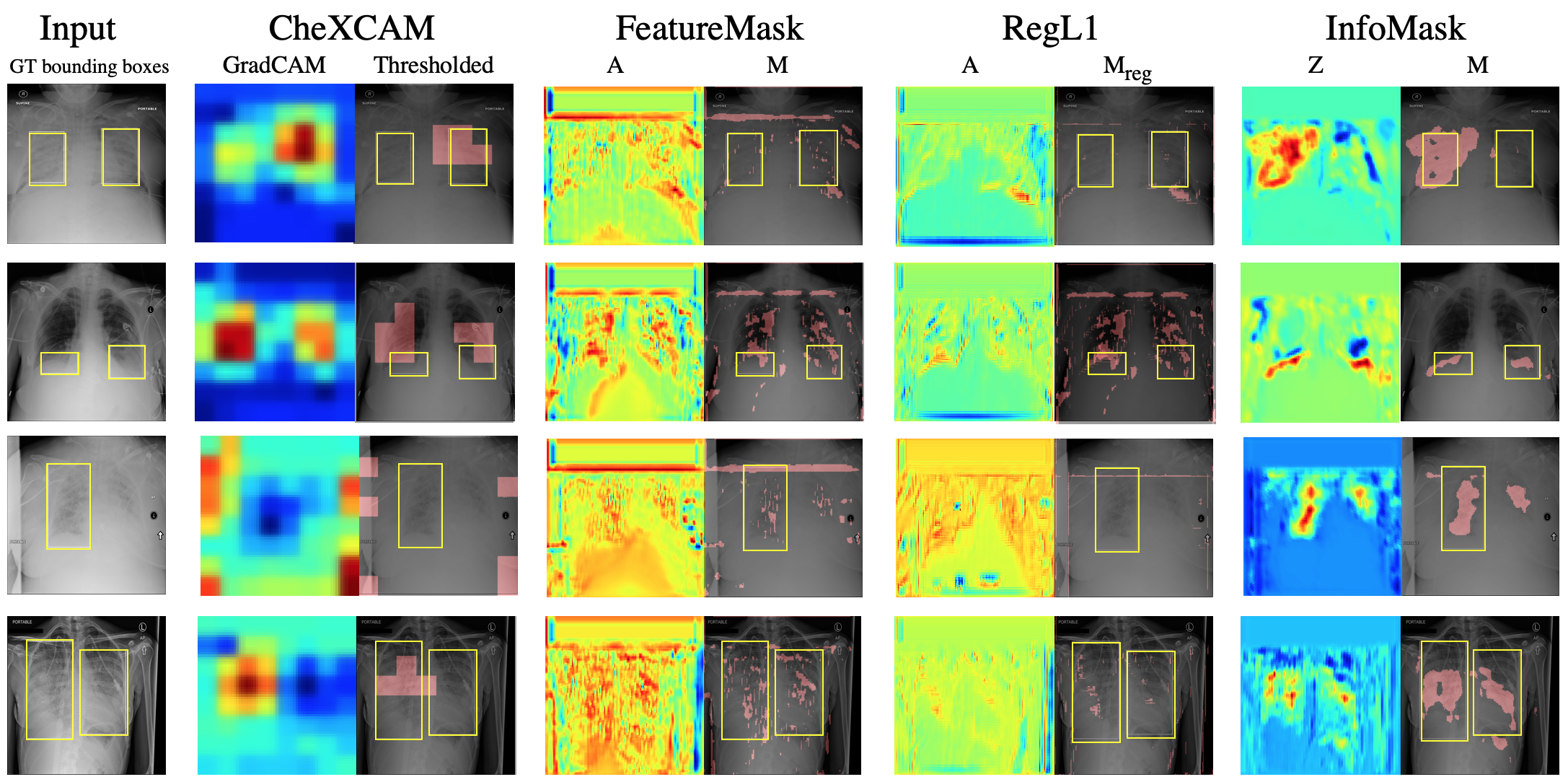}
  \caption{Examples of pneumonia localization for  various methods. GT bounding boxes shown in yellow.}
  \label{fig-att}
\end{figure}

\begin{figure}[h!]
\centering
  \includegraphics[width=.9\textwidth]{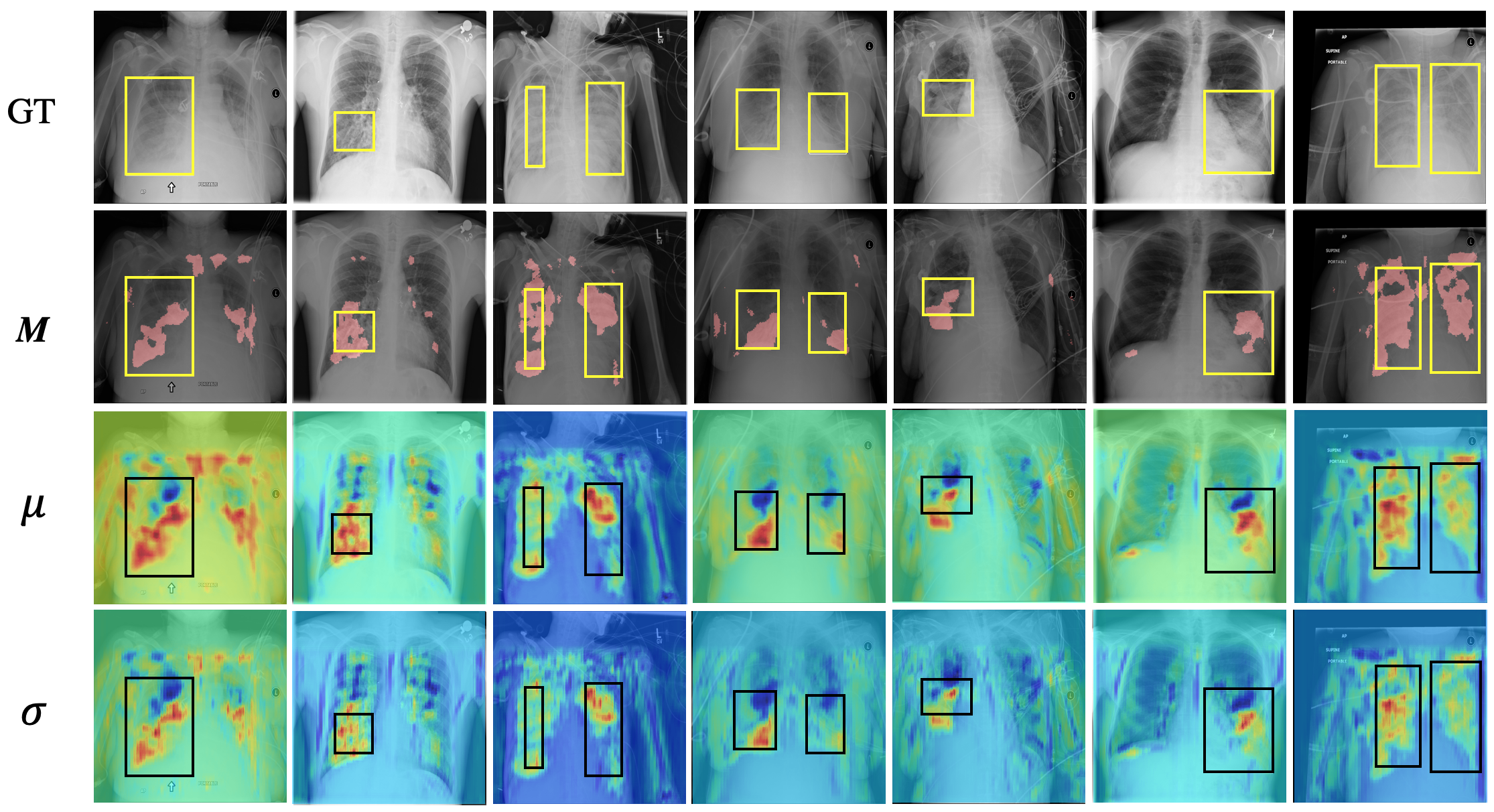}
  \caption{A few localization samples of InfoMask with mean ($\mu$) and variance ($\sigma$) maps.}
  \label{varianc-fig}
\end{figure}

We then compare our proposed InfoMask to four competing disease localization methods:  (i) GradCAM, gradient-weighted class activation mapping + baseline, i.e., during inference, we replace $\bm{M}$ in Fig~\ref{diag1} with GradCAM; (ii) FeatureMask, masking the latent representation without KL divergence regularization + baseline; (iii) RegL1, L1 regularization over the generated masks instead of KL regularization; (iv) CheXCAM, GradCAM applied to the last layer of CheXNet~\cite{rajpurkar2017chexnet}. Even though each patient could have multiple disease classes at the same time, we focus only on pneumonia disease detection (vs. normal) to analyze whether our method is able to only localize target regions in a complex environment where other diseases might also be present. Note that the results(Table~\ref{table1} and Figures~\ref{fig:kde},~\ref{fig-att}, and ~\ref{varianc-fig}) reported next are based on the thresholded masks using the best threshold value, i.e., optimized, for each method, to minimize localization error over the validation set. To select the best epoch based on a validation set we first select $N$ checkpoints which produce highest classification accuracy and then select the epoch with the highest localization score among them.
As the detected thresholded masks could potentially have largely diverse patterns (e.g., from sparse disjoint localizations scattered over the whole image to large connected components, to anything in between), computing a single representative bounding box, as is provided by ground truth bounding box annotations, is not straightforward. 
\vspace*{-7mm}
\begin{table}[]
\setul{}{.8pt}
\centering
\caption{Disease localization performance evaluation of the proposed InfoMask vs. competing methods. IoP, FPR, and FNR represent the localization performance while Acc. and AUC show the classification performance.} 
\label{table1}
\setlength{\tabcolsep}{5pt}
\begin{tabular}{lccccc}
\hline
 & IoP & FPR & FNR & Acc. & AUC \\ \hline
GradCAM & 0.12 $\pm$ 3.0e-04 & 0.196 $\pm$ 2.0e-04 &  0.20 $\pm$ 2.0e-04 & $0.8221$ & 0.8333\\
FeatureMask & 0.19 $\pm$ 2.0e-04 & 0.095 $\pm$ 4.7e-05 & 0.81 $\pm$ 2.0e-04 & 0.8236 & 0.8375\\
RegL1 & 0.11 $\pm$ 2.0e-04 &  0.010 $\pm$ 6.5e-06 & 0.99 $\pm$ 3.7e-05 & 0.8170 & 0.8306\\
CheXCAM & 0.34 $\pm$ 5.0e-04 & 0.077 $\pm$ 7.0e-05 & 0.71 $\pm$ 4.0e-04 &  0.8400 &  0.8644\\
InfoMask &  0.44 $\pm$ 5.0e-04 & 0.025 $\pm$ 3.6e-05 & 0.80 $\pm$ 3.0e-04 & 0.8248 & 0.8251\\ \hline
\end{tabular}
\end{table}
\vspace*{-5mm}
Therefore, we replace the intersection over union (IoU) quality metric, commonly used for evaluating bounding box predictions, with a proposed intersection over predicted area  (IoP), which reflects what percentage of the predicted area is inside the ground truth bounding box. As a small predicted areas inside the box can lead to a high score, we also compute false positive and negative rates, FPR and FNR respectively to measure over- and under-predicted areas. 
As reported in Table~\ref{table1}, the proposed InfoMask outperforms the competing methods by a large margin on IoP (at least 10\% better), and obtains the second best FPR (only 1.5\% higher than the lowest FPR). Examining the FPR values, it can be inferred that GradCAM tends to highlight larger areas of the input outside of the ground truth bounding boxes. 
\vspace*{-5mm}
\vspace*{-5mm}
Form the FNR column, we note that RegL1 generates smaller areas inside the boxes. Although the focus of the current study is not to improve classification accuracy, our proposed method achieves only slightly smaller classification accuracy ($<2\%$) but with only 10\% (7,000,000 vs. 700,000) of the parameters of CheXNet. The kernel density estimation plots in Figure~\ref{fig:kde} support the quantitative results for the test images. Note how InfoMask obtains higher densities at larger IoP values (note: green curve in (a) for IoP$\in[0.5,1]$), smaller FPR density (green peak in (b) for FPR$\in[0,0.1]$) and in second best (behind CheXCAM) for FNR values. 
For a better interpretation of Table~\ref{table1}, we visualized a few samples of the attention maps and masked ones in Figure~\ref{fig-att} along with the ground truth (GT) bounding boxes in yellow. 
\vspace*{-3mm}
\vspace*{-7mm}
In Figure~\ref{varianc-fig}, we visualized a few mean and variance samples computed for test images. As shown, there is less variance in the areas where the model is confident about absence of disease signs. As visualized InfoMask was able to localize pneumonia from images with different intensity distributions without using any bounding-box level annotation.
\vspace*{-5mm}
\vspace*{-5mm}
As can be seen, FeatureMask and RegL1 produce scattered attention maps that cover only small portions of the GT bounding boxes. Among all, the proposed InfoMask generates contiguous  attention areas with most  agreement with ground truth boxes. 

\section{Conclusion}
We proposed InfoMask, a method to localize disease-discriminatory regions trained with only image-level labels. Owing to the regularized variational latent representation with an attention mechanism, InfoMask generate contiguous and focused localization masks with higher agreement with ground truth annotations than competing methods (e.g., widely used GradCAM) without resorting to any bounding-box level annotations. A direction for future work aims at improving both classification and localization objectives by using stronger classification backbone models. 

\section*{Acknowledgement}
We thank Dr. Joseph Paul Cohen for his insightful discussions and comments.

%

\bibliographystyle{splncs04}
\bibliography{biblo}
\end{document}